\documentclass{article}

 \usepackage[preprint]{tackling_climate_workshop_style}

\usepackage[utf8]{inputenc} 
\usepackage[T1]{fontenc}    
\usepackage{hyperref}       
\usepackage{url}            
\usepackage{booktabs}       
\usepackage{amsfonts}       
\usepackage{nicefrac}       
\usepackage{microtype}      
\usepackage{amsmath}
\usepackage{algorithmic}
\usepackage{array}
\usepackage[tight,footnotesize]{subfigure}
\usepackage{url}
\usepackage{accents}
\usepackage{booktabs}
\usepackage[table,xcdraw]{xcolor}
\usepackage[pdftex]{graphicx}

\title{\textit{Dynamic weights enabled Physics-Informed Neural Network for simulating the mobility of Engineered Nano-particles in a contaminated aquifer }} 

\author{
  Shikhar Nilabh,  Fidel Grandia \\
   Amphos 21 Consulting S.L.,\\
  Barcelona, Spain  \\
  \texttt{\{shikhar.nilabh, fidel.grandia\}@amphos21.com} \\
}

\begin{document}

\maketitle

\begin{abstract}
Numerous polluted groundwater sites across the globe require an active remediation strategy to restore natural environmental conditions and local ecosystem. The Engineered Nano-particles (ENPs) have emerged as an efficient reactive agent for the in-situ degradation of groundwater contaminants. While the performance of these ENPs has been highly promising on the laboratory scale, their application in real field case conditions is still limited. The complex transport and retention mechanisms of ENPs hinder the development of an efficient remediation strategy. Therefore, a predictive tool for understanding the transport and retention behavior of ENPs is highly required. The existing tools in the literature are dominated with numerical simulators, which have limited flexibility and accuracy in the presence of sparse datasets. This work uses a dynamic, weight-enabled Physics-Informed Neural Network (dw-PINN) framework to model the nano-particle behavior within an aquifer. The result from the forward model demonstrates the effective capability of dw-PINN in accurately predicting the ENPs mobility. The model verification step shows that the relative mean square error (MSE) of the predicted ENPs concentration using dw-PINN converges to a minimum value of $1.3{e^{-5}}$. In the subsequent step, the result from the inverse model estimates the governing parameters of ENPs mobility with reasonable accuracy. The research demonstrates the tool's capability to provide predictive insights for developing an efficient groundwater remediation strategy.

\end{abstract}

\section{Introduction}

 The water scarcity is a rapidly growing concern with more than a billion of people deprived of safe drinking water[1]. The widespread presence of contamination in aquifers poses a great threat to the human health, environmental quality and socioeconomic development[2], [3]. Furthermore, the groundwater quality is intricately linked to the climate change. On one hand, the persistence of groundwater pollution can lead to the degradation of soil quality[4], forestry[5] and even coastal ecosystem[6]; thus accelerating the climate change processes. Conversely, the climate change can intensify the fresh water demands, and enhance sea water intrusion[7]. With the escalating industrial activities in the past decades, implementation of groundwater remediation techniques has become critical in controlling the water quality.
 
 Several remediation techniques have been developed in the literature for the restoration of groundwater resources[8]–[10]. In the recent years, injection of ENPs in a contaminated aquifer has proven to be highly efficient in the groundwater remediation[11], [12]. While extensive research has been done in developing and testing these ENPs at laboratory scale, their field scale injection for groundwater remediation is still limited[10]. The limitation of  ENPs application is attributed to its highly complex transport and retention behavior which is often difficult to predict. While several numerical and data-driven tools have been proposed in the literature for predicting the behavior of ENPs, their application is often challenged by data sparsity and heterogeneity in hydrogeological parameters[13]. A Physics-Informed Neural Network (PINN) can overcome these challenges by embedding the knowledge of governing equations into the learning method of a neural network[14], [15]. However, the application of PINNs in the subsurface flow and transport is still rare and have been only limited to contaminant mobility[13], [15]. To our best of knowledge, this paper is the first to implement Physics-Informed Neural Network for simulation of ENPs mobility in a saturated sand. 
 
 While there are several advantages of using PINN model, the gradient imbalance problem reduces its accuracy[16], [17]. To overcome this problem, [13] proposed dynamic weight strategy for PINN (dw-PINN) where PINN is modified by assigning dynamically updated weights for the balance of   loss function. The dynamic weight adaption has been highly effective in solving flow and transport related equations and therefore in this research work, the model is developed with dw-PINN. With this method, this research study aims to provide a robust predictive tool for bridging the gap between laboratory testing of ENPs and its field implementation. In this regard, a two-fold objective of this work is defined: a) development and verification of forward model simulating nano-particle's transport and retention in a 1-Dimensional column-filled sand. b) inverse modeling using dw-PINN for estimating the hydrogeological properties and physiochemical properties of ENPs. Here, the overall research goal is limited to demonstrating the tool’s capability in understanding ENPs mobility, and its application in a real field case scenario is beyond the scope of this work.

\section{  Methodology}
The nano-particle's fate in the aquifer is governed by transport due to groundwater flow and retention due to particle-sand interaction. Equation (1) and (2) represent the modified Advection Dispersion equation for the ENPs transport and retention in a 1D column-filled saturated sand representing a small-scale aquifer[18]. Equation 3 represents the pulse of ENPs 1 $kg\times m^{-3}$ assigned at the boundary $c(0,t)$.The other boundary, $c(1,t)$, is assigned with a Neumann boundary condition. The initial concentration of ENPs, $c(x,0)$ is considered to be zero.

\begin{align}
	\frac{\partial(\theta c)}{\partial t}+\frac{\partial(s)}{\partial t}=-\nabla(vc)+\nabla({{(D}_e+\ \propto}_Lv)\nabla c)
\end{align}
\begin{align}
	\frac{\partial(s)}{\partial t}=\theta k_ac-k_ds                                                                                                                       
\end{align}
\begin{align}
	c\left(0,t\right)=\left(\frac{1}{1+e^{-0.02\left(t-500\right)}}\right)\times\left(\frac{1}{1+e^{0.02\left(t-4100\right)}}\right)
\end{align}
Where $\theta$ is the porosity, $D_e$ is molecular dispersivity $(m^2\times s^{-1})$, ${\ \propto}_L$ is dispersivity $(m)$, $k_a (s^{-1})$and  $k_d$ $(s^{-1})$ is the attachment and the detachment coefficient of ENPs respectively, $c (kg\times m^{-3})$ and $s (kg\times m^{-3})$ are aqueous and retained concentration of ENPs respectively.

Two fully connected neural networks are considered to approximate the aqueous and retained concentration of ENPs (Figure 1). Each of the network consists of 6 hidden layers each with 50 neurons. 3000 collocation points are used for enforcing boundary conditions whereas, with Lattice hypercube Sampling, 15000 collocation points are selected for enforcing equation 1 and 2. Sigmoid activation function is used for incorporating non-linearity in the neural network. The model is run for 20000 steps of gradient descent-ascent process using Adam optimizer[19].
\begin{figure*} [ht!]
	\vspace{2mm}
	\centering
	\includegraphics[width=\linewidth]{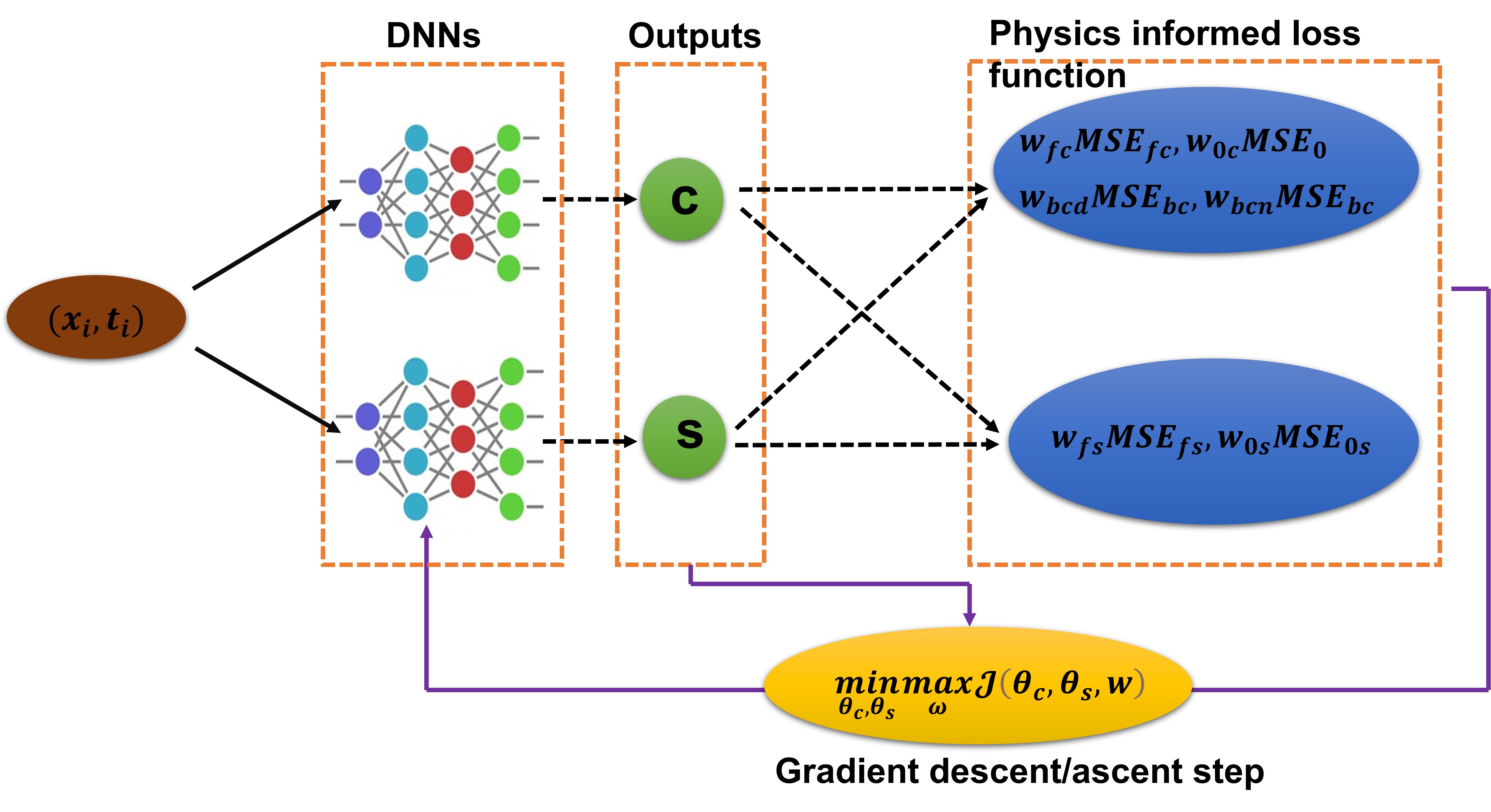}
	\caption{Flowchart of dw-PINN framework representing two Deep Neural Networks with time and spatial points as input and concentration of aqueous and retained ENPs as output.  The loss function $\mathcal{J}$\ consists of 6 Mean square error term MSEs corresponding to the boundary and governing PDE equations. For each MSE, a weight term is assigned to balance the loss function. }
	\label{Fig1}
	\vspace{-4mm}
\end{figure*}

The model is developed in two stages. In the first stage, a forward dw-PINN model is developed and verified with a benchmark model. This benchmark model simulates ENPs injection in a column filled sand with porosity of 0.2, dispersivity of 0.05 $m$, attachment rate of 0.001 $s^{-1}$ and detachment rate of 0.0001 $s^{-1}$. For this simulation, a Finite Element Method (FEM) based software Comsol Multiphsysics is used. In the second stage, an inverse dw-PINN model is developed to estimate the hydrogeological parameters with the tracer simulation and physiochemical parameters of ENPs. For the parameter estimation, dw-PINN model requires a training dataset which is generated using Comsol.

\section{Results and Discussion }
\subsection{Model verification}
In the first stage, a forward model is generated with dw-PINN and compared with the results of Comsol model. Figure 2 (a) and 2 (b) show a close agreement between the result of dw-PINN model and Comsol simulation for breakthrough curve and retention profile respectively. The calculated $R^2$ value of the plots in figure 2 (a) and 2 (b) is 0.986 and 0.993 respectively. The breakthrough curve in figure 2 (a) shows that the result of   dw-PINN model is relatively more smooth in the region where Comsol model has undergone overshooting and undershooting. Overall, result demonstrates the effective capability of dw-PINN in predicting the mobility  and retention of ENPs in saturated sand.  
\begin{figure*} [ht!]
	\vspace{2mm}
	\centering
	\includegraphics[width=\linewidth]{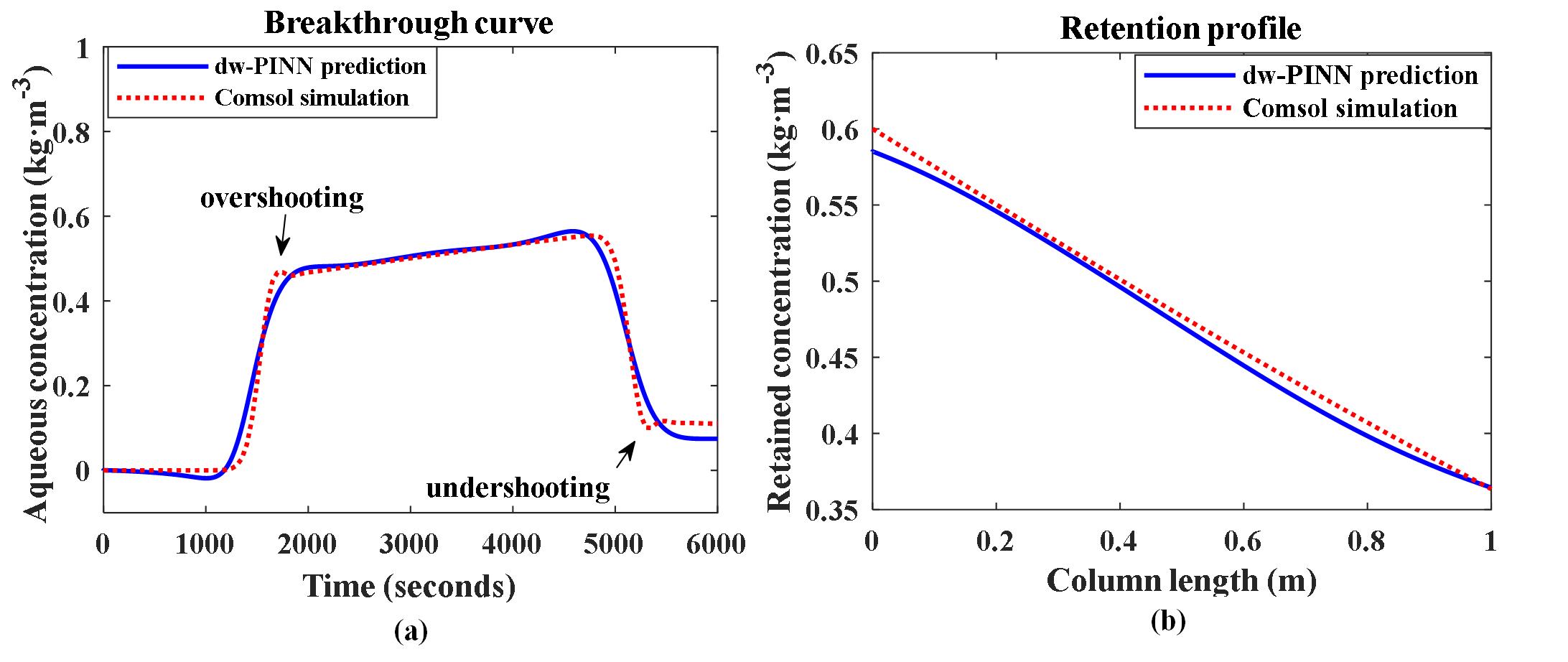}
	\caption{ Result comparison of dw-PINN and Comsol model; (a) breakthrough curve (b) retention profile }
	\label{Fig2}
	\vspace{-4mm}
\end{figure*}

\subsection{Inverse modeling}
In the subsequent step, the inverse model is developed to estimate   the hydrogeological and ENP's properties. Figure 3 (a) and 3 (b) shows the breakthrough curve for the tracer and ENPs simulated using dw-PINN tool and its close fit with the training dataset generated using Comsol. The result highlights model’s ability to accurately simulate the transport dynamics of the tracer and ENPs.

\begin{figure*} [ht!]
	\vspace{2mm}
	\centering
	\includegraphics[width=\linewidth]{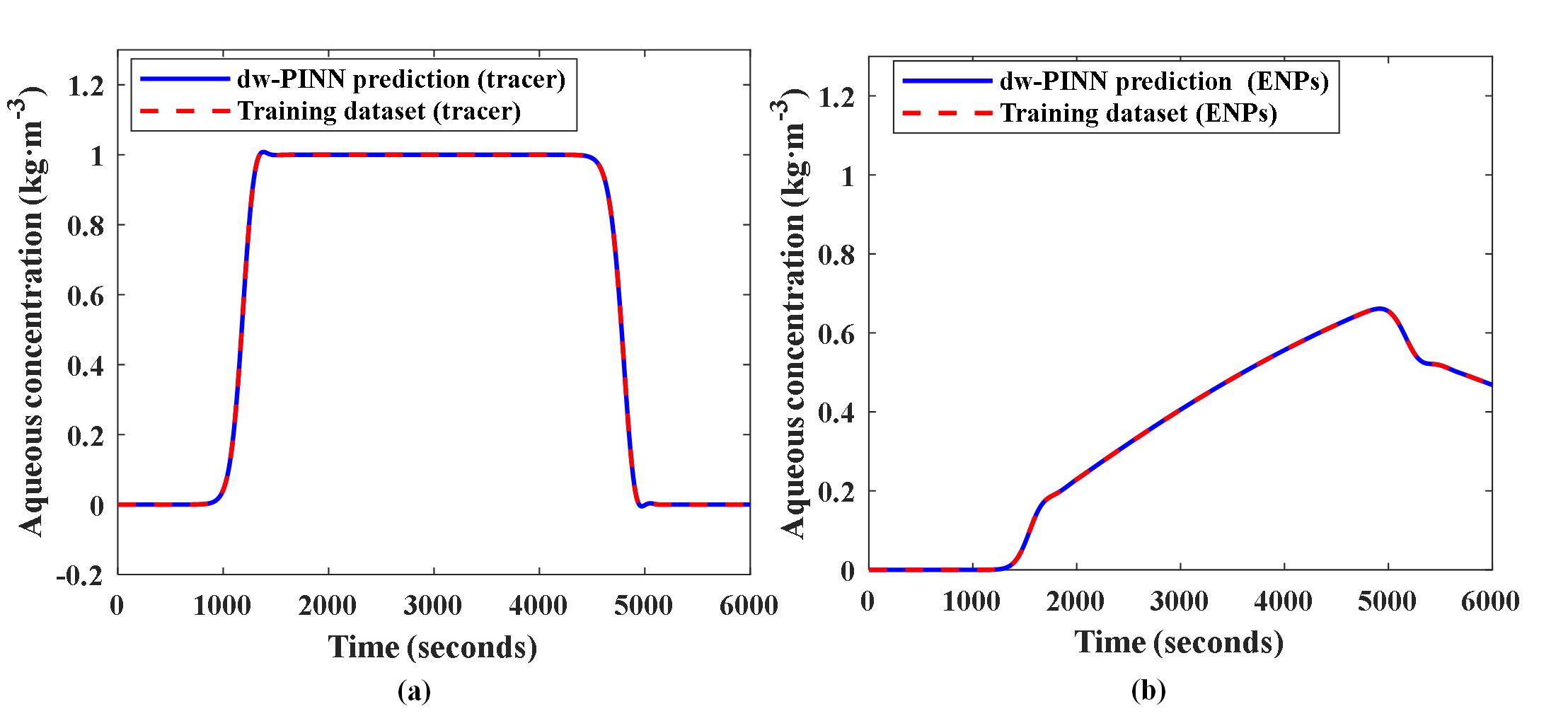}
	\caption{ Result comparison of inverse dw-PINN model and the training dataset generated using Comsol; (a) breakthrough curve for tracer (b) breakthrough curve for ENPs }
	\label{Fig3}
	\vspace{-4mm}
\end{figure*}

The sand and ENPs properties estimated using dw-PINN model are listed in table \ref{tab:1}. The table \ref{tab:1} shows a close match in the estimated value of the governing parameters with the real values on which training dataset is generated. The error has been relatively higher for the estimated value of dispersivity and detachment coefficient. This could be attributed to the non-uniform sensitivity of the ENPs concentration with respect to governing parameters.

\begin{table}[h]
		\centering
	\caption{The model estimation for the governing parameters is compared with their real values on which the model is trained}
	\vspace{3mm}
	\label{tab:1}
	\begin{tabular}{|l|l|l|l|l|}
		\hline
		& Porosity & \multicolumn{1}{c|}{\begin{tabular}[c]{@{}c@{}}Dispersivity \\  (m)\end{tabular}} & \begin{tabular}[c]{@{}l@{}}Attachment \\ coefficient   $(s^{-1})$ \end{tabular} & \begin{tabular}[c]{@{}l@{}}Detatchment   \\ coefficient $(s^{-1})$\end{tabular} \\ \hline
		Real values        & 0.2      & 0.005                                                        & 0.0007                                                                    & 0.0001                                                                     \\ \hline
		Estimated   values & 0.201    & 0.0059                                                       & 0.00071                                                                   & 0.00018                                                                    \\ \hline
	\end{tabular}
\end{table}

\section{Conclusion}

The research work intends to develop a robust modeling tool for providing relevant predictive insights in the development of groundwater remediation plan. The result demonstrates An effective forward and inverse dw-PINN based tool for studying the nano-particle's transport and retention behavior in a small-scale aquifer. The forward model has been verified using the results from Comsol Multiphysics with the loss function converges to a minimum value of $1.3e^{-5}$. The inverse model based on the same verified framework estimates the governing parameters for ENPs with a maximum accuracy of $98.6\%$. The tool can be instrumental in bridging the existing gap in the laboratory-scale study and field-scale implementation. This dw-PINN tool can be used in a sequential manner for developing an efficient groundwater remediation plan using ENPs. In the first step, the inverse model can be used for estimation of governing parameters based on the experimental data of ENPs injection in a small-scale aquifer. In the second step, a forward model could be developed based on the estimated parameters to provide prediction insights relevant for groundwater remediation.

\section*{References}

\medskip

\small

[1]	N. R. Ochilova, G. S. Muratova, y D. R. Karshieva, «The Importance of Water Quality and Quantity in Strengthening the Health and Living Conditions of the Population», CENTRAL ASIAN JOURNAL OF MEDICAL AND NATURAL SCIENCES, vol. 2, n.o 5, pp. 399–402, 2021.

[2]	M. D. Einarson y D. M. Mackay, «Peer reviewed: predicting impacts of groundwater contamination». ACS Publications, 2001.

[3]	A. Fares, Emerging issues in groundwater resources. Springer, 2016.

[4]	S. D. Keesstra et al., «Soil as a filter for groundwater quality», Current Opinion in Environmental Sustainability, vol. 4, n.o 5, pp. 507–516, 2012.

[5]	R. C. Simmons, A. J. Gold, y P. M. Groffman, «Nitrate dynamics in riparian forests: groundwater studies», Wiley Online Library, 1992.

[6]	L. M. Hernández-Terrones, K. A. Null, D. Ortega-Camacho, y A. Paytan, «Water quality assessment in the Mexican Caribbean: impacts on the coastal ecosystem», Continental Shelf Research, vol. 102, pp. 62–72, 2015.

[7]	M. M. Sherif y V. P. Singh, «Effect of climate change on sea water intrusion in coastal aquifers», Hydrol. Process., vol. 13, n.o 8, pp. 1277-1287, jun. 1999, doi: 10.1002/(SICI)1099-1085(19990615)13:8<1277::AID-HYP765>3.0.CO;2-W.

[8]	P. Lojkasek-Lima et al., «Evaluating TCE abiotic and biotic degradation pathways in a permeable reactive barrier using compound specific isotope analysis», Groundwater Monitoring  Remediation, vol. 32, n.o 4, pp. 53–62, 2012.

[9]	E. A. Seagren, B. E. Rittmann, y A. J. Valocchi, «Quantitative evaluation of the enhancement of NAPL-pool dissolution by flushing and biodegradation», Environmental science \ technology, vol. 28, n.o 5, pp. 833–839, 1994.

[10]	T. Zhang et al., «In situ remediation of subsurface contamination: opportunities and challenges for nanotechnology and advanced materials», Environmental Science: Nano, vol. 6, n.o 5, pp. 1283–1302, 2019.

[11]	T. Tosco, M. P. Papini, C. C. Viggi, y R. Sethi, «Nanoscale zerovalent iron particles for groundwater remediation: a review», Journal of cleaner production, vol. 77, pp. 10–21, 2014.

[12]	C. M. Kocur, D. M. O’Carroll, y B. E. Sleep, «Impact of nZVI stability on mobility in porous media», Journal of contaminant hydrology, vol. 145, pp. 17–25, 2013.

[13]	Q. He, D. Barajas-Solano, G. Tartakovsky, y A. M. Tartakovsky, «Physics-informed neural networks for multiphysics data assimilation with application to subsurface transport», Advances in Water Resources, vol. 141, p. 103610, 2020.

[14]	M. Raissi, P. Perdikaris, y G. E. Karniadakis, «Physics-informed neural networks: A deep learning framework for solving forward and inverse problems involving nonlinear partial differential equations», Journal of Computational physics, vol. 378, pp. 686–707, 2019.

[15]	Q. He y A. M. Tartakovsky, «Physics-Informed neural network method for forward and backward advection-dispersion equations», Water Resources Research, vol. 57, n.o 7, p. e2020WR029479, 2021.

[16]	L. McClenny y U. Braga-Neto, «Self-Adaptive Physics-Informed Neural Networks», Available at SSRN 4086448.

[17]	S. Li y X. Feng, «Dynamic Weight Strategy of Physics-Informed Neural Networks for the 2D Navier–Stokes Equations», Entropy, vol. 24, n.o 9, p. 1254, 2022.

[18]	I. L. Molnar, W. P. Johnson, J. I. Gerhard, C. S. Willson, y D. M. O’Carroll, «Predicting colloid transport through saturated porous media: A critical review», Water Resources Research, vol. 51, n.o 9, pp. 6804–6845, 2015.

[19]	D. P. Kingma y J. Ba, «Adam: A method for stochastic optimization», arXiv preprint arXiv:1412.6980, 2014.

%
%

\end{document}